\documentclass[journal]{IEEEtran}
\usepackage{cite}

\usepackage{subfigure}
\usepackage{graphicx}
\usepackage{graphics}
\usepackage{amssymb}
\usepackage{color}
\usepackage{booktabs}
\usepackage{threeparttable}

\ifCLASSINFOpdf
\else
\fi
\usepackage[cmex10]{amsmath}
\usepackage{array}
\usepackage{url}

\usepackage[switch,left]{lineno}
\usepackage{xcolor}

\begin{document}
%
\title{Refining time-space traffic diagrams: A simple multiple linear regression model}

%
%
%

\author{Zhengbing~He, {\it Senior Member, IEEE}
\thanks{
Z. He is with Senseable City Lab, Massachusetts Institute of Technology, Cambridge MA, United States ({\it he.zb@hotmail.com}). 
Corresponding author: Zhengbing He.
}
\thanks{\textcolor{blue}{Accepted in Sep. 2023, and to be published with DOI: 10.1109/TITS.2023.3316593}}}



%
%

\markboth{Intelligent Transportation Systems, IEEE Transactions on}%
{}
%



\maketitle

\begin{abstract}
A time-space (TS) traffic diagram, which presents traffic states in time-space cells with color, is an important traffic analysis and visualization tool. Despite its importance for transportation research and engineering, most TS diagrams that have already existed or are being produced are too coarse to exhibit detailed traffic dynamics due to the difficulty of collecting high-fidelity traffic data. To increase the resolution of a TS diagram and enable it to present ample traffic details, this paper introduces the TS diagram refinement problem and proposes a multiple linear regression-based solution. Data collected at different times, in different locations and even in different countries are employed to thoroughly evaluate the accuracy and transferability of the proposed model. Two tests, which attempt to increase the resolution of a TS diagram 4 and 16 times, are carried out to evaluate the performance of the proposed model. In the increase-4-times test, the errors represented by Mean Absolute Percentage Error are all less than 0.1, and in the increase-16-times test all less than 0.17. Model comparison demonstrates that the proposed model outperforms the classic adaptive smoothing method in refining TS diagrams. All the strict tests with diverse data show that the proposed model, despite its simplicity, is able to refine a TS diagram with promising accuracy and reliable transferability. The proposed refinement model will ``save" widely existing TS diagrams from their blurry ``faces" and enable TS diagrams to show more traffic details.

\end{abstract}


%
\IEEEpeerreviewmaketitle

\section{Introduction}\label{sec:Intro}

\IEEEPARstart{A}{} time-space (TS) traffic diagram, in which the x-axis indicates time, the y-axis denotes space, and a diverse array of colors represents different traffic states, is an important traffic analysis and visualization tool \cite{Zheng2010,Zheng2011b,Lu2017,Grumert2018,Wang2022}.
From a TS diagram, we can identify traffic bottlenecks \cite{Chen2004b,Ban2008}, understand traffic characteristics \cite{Wan2020,He2015}, predict travel time \cite{Dai2017,Zhang2017c}, and even estimate traffic emissions \cite{He2020}. 
Almost all traffic data centers and display platforms, such as California's Performance Measurement System (PeMS) in the United States, employ TS diagrams to visualize traffic dynamics.

A typical method of constructing a TS diagram is to virtually partition a TS plane into homogeneous cells and then fill the cells with colors that are determined based on the traffic states of the cells.
For a TS diagram, the average traffic speed in a TS cell is commonly used to represent the traffic state in practice.
The main reasons of using speed instead of density and flow are as follows:
(i) Speed could be directly measured by stationary detectors \cite{Treiber2002} or floating cars \cite{He2017};
(ii) Speed is more straightforward for both engineers and travelers;
(iii) One could easily estimate the travel time of an individual vehicle from the TS diagram of traffic speed \cite{Zhang2017c,He2019}.

\begin{figure}[h]
    \centering
    \includegraphics[width=3.5in]{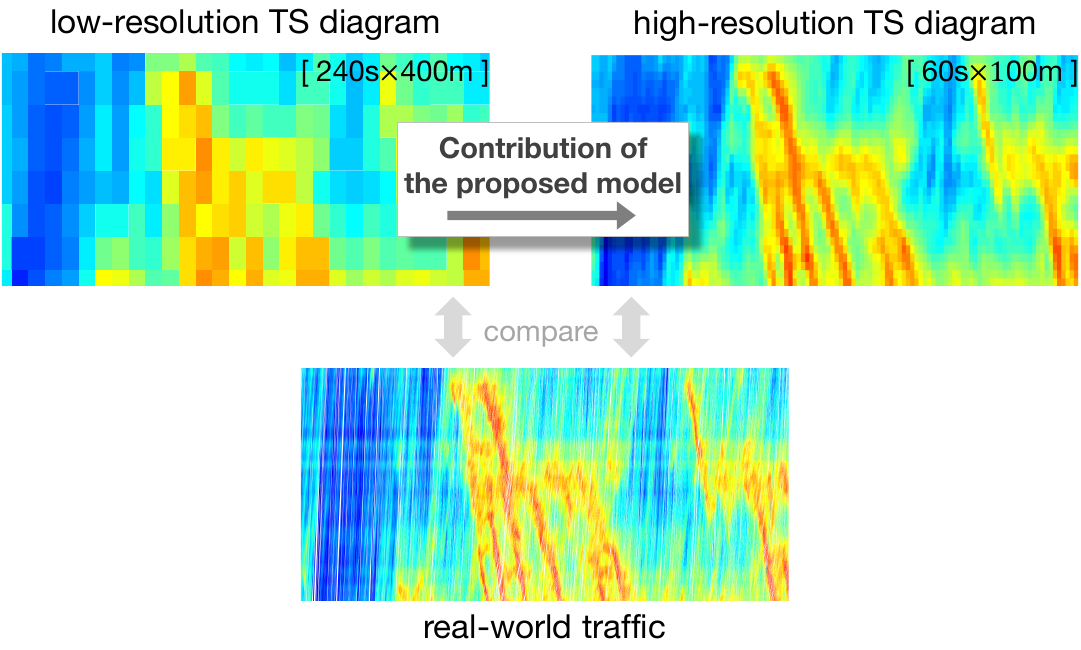}
    \caption{Comparison of low- and high-resolution TS diagrams.}
    \label{fig:ResolutionComparison}
\end{figure}

The key step of constructing a TS diagram is to measure the speed that corresponds to each TS cell.
Ideally, traffic speed can be accurately calculated by using 100\% high-fidelity vehicle trajectories that pass through TS cells, and a high-resolution TS diagram with small cells can be constructed to exhibit detailed traffic and analyze traffic characteristics such as stop-and-go waves and wave directions (Figure~\ref{fig:ResolutionComparison}).
Unfortunately, existing information technology does not support the wide collection of such 100\% high-fidelity vehicle trajectory data in practice, and only sampled traffic data could be collected for daily applications.

From sampled data traffic state estimation (TSE) is usually employed to derive missing speed information. 
For example, given the loop detector data at two spatial locations, the intermediate traffic states can be estimated based on traffic characteristics \cite{Treiber2002,Treiber2011a}, the Kalman filter \cite{Wang2005,Wang2007}, the Newell's three-detector approach \cite{Laval2012,Deng2013}, etc.
Given floating car data, traffic states can be approximated by mapping the data to carefully selected spatial cells that overlap a freeway \cite{He2017,He2017a}.
However, the loop detector density cannot be as high as we expected due to financial limitations \cite{Seo2017}. For example, the detector spacing is usually between several hundred meters and several kilometers.
The interval that aggregates the uploaded loop detector data cannot be too short owing to the limitation of transmission bandwidth and storage space; it is common to set the time interval of uploading detector data from 1 min to 10 min.
For similar reasons, the sampling and penetration rates of floating car data are usually low in practice.
It is quite typical for a floating car to upload its driving status data every 1 minute or even longer. 
The sparse nature of TS sampled data makes it difficult to construct a high-resolution TS diagram even with the aid of advanced TSE methods.

To extract more traffic details from a widely-existed low-resolution TS diagram, a meaningful and interesting attempt is to directly refine a TS diagram by increasing its resolution.
In computer science, this approach is referred to as {\it superresolution}, i.e., enhancing the resolution of an imaging system \cite{Tsai1984,Irani1990}.
Superresolution is an active field, particularly with the development of deep-learning approaches \cite{Wang2015,Dong2016,Zhang2018}.
Convolutional neural networks (CNNs) are the most popular foundation of various deep learning-based superresolution approaches owing to their powerful capability of processing images \cite{Wu2021}. 
Machine learning-based superresolution approaches do not require solving complex equations, while the drawbacks of these approaches are twofold.
First, the computational cost is usually high, hindering the real-time applications of superresolution approaches \cite{Wu2021}.
Second, a large number of images are needed for training deep-learning models \cite{Zhang2022}.
For the purpose of refining a TS diagram in transportation, the second drawback is a major barrier, i.e., it is unrealistic to collect high-fidelity and high-sampling-rate traffic data as we previously mentioned, and thus, we do not have enough target images to train the deep-learning models.
Although one may consider traffic simulation as an option for generating samples for training \cite{Shi2022}, it is not clear if a simulation, which is usually unable to well reproduce human drivers' irrational driving behaviors, could produce qualified samples.

\begin{figure}[h]
    \centering
    \includegraphics[width=3.5in]{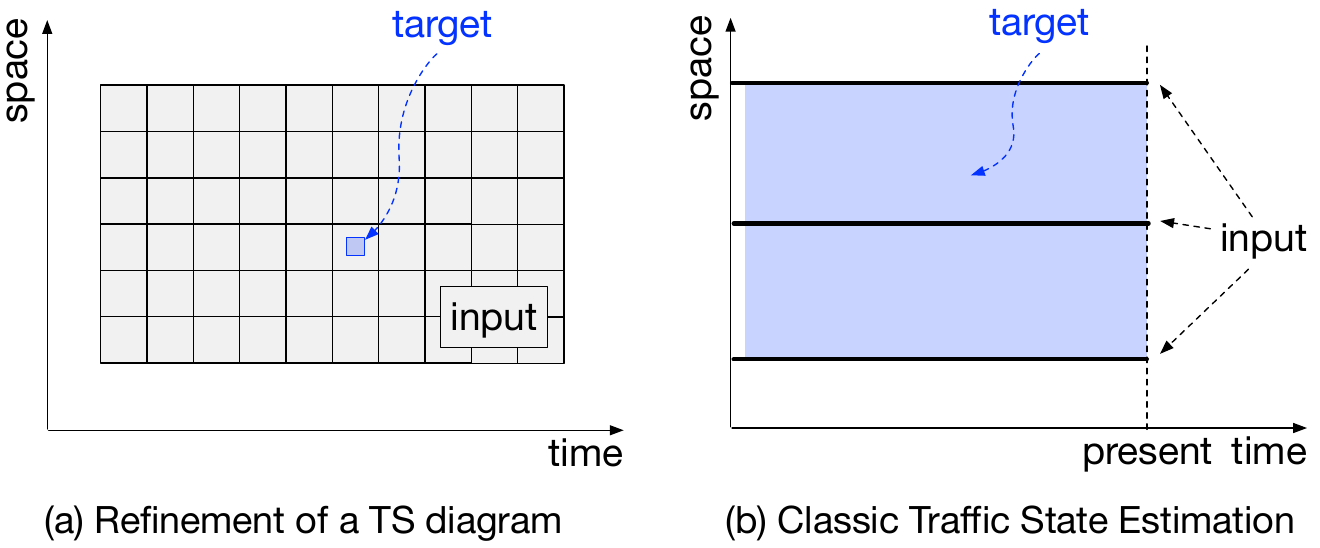}
    \caption{Comparison of the refinement of a TS diagram with the traffic state estimation.}
    \label{fig:Comparison}
\end{figure}

It is worth noting that the problem faced is different from the traditional TSE problem, although the refinement of a TS diagram is to estimate traffic states in essence.
Referring to Figure~\ref{fig:Comparison}, the target of estimation (i.e., the output) in refining a TS diagram is the traffic speed in a smaller TS cell, while the target in the TSE problem is usually the traffic states in between spatially-located detectors. 
The inputs in refining a TS diagram are spatiotemporally-discrete traffic states (represented by only traffic speed).
In particular, the inputs even include the traffic states in future, i.e., the states contained by the cells on the right-hand side of the targeted cell.
In contrast, the inputs of the TSE problem are traffic states (usually represented by three traffic variables, i.e., speed, density, and flow) in two separated locations, and it is common to assume that we do not know the traffic states in future.
In addition, the estimation resolution (represented by the spatial length and time width of a pixel or cell) of the TSE methods that are commonly proposed based on the traffic flow characteristics has to consider the speed of traffic wave propagation, making the resolution cannot be determined flexibly and the time width is often too small compared to the spatial length for practical purposes \cite{Seo2022}.

In summary, the resolution of most TS diagrams that have already existed or are being generated is quite coarse for detailed analyses and applications, mainly owing to the limitation of the existing information technology and traffic infrastructure investment.
A method that could directly increase the resolution of TS diagrams with no use of a large number of high-resolution TS diagrams for model training is urgently needed, so that the widely-existed TS diagrams could exhibit more traffic details and better help analyze traffic dynamics.
To fill this gap, this paper proposes a multiple linear regression-based model to refine a low-resolution TS diagram by directly increasing its resolution.
To the best of our knowledge, this is the first work that introduces the refinement problem of a TS diagram and successfully solves it with only a small amount of high-fidelity traffic data. 
Note that we cannot quantify the ``low" resolution of a TS diagram that surely needs the proposed method. 
Instead, the proposed method could make the existing TS diagrams clearer and show more details in exhibiting traffic dynamics. 

The remainder of the paper is organized as follows:
Section \ref{sec:Model} introduces the refinement problem of a TS diagram and proposes a refinement model based on multiple linear regression.
Section \ref{sec:Evaluation} evaluates the proposed refinement model using three completely different datasets in two strict tests.
Section \ref{sec:Comparison} compares the proposed model with one of the classic TSE models.
The discussion and conclusion are presented in Sections \ref{sec:Discussion} and \ref{sec:Conclusion}, respectively.

\section{Model}\label{sec:Model}

\begin{figure}[h]
    \centering
    \includegraphics[width=2.7in]{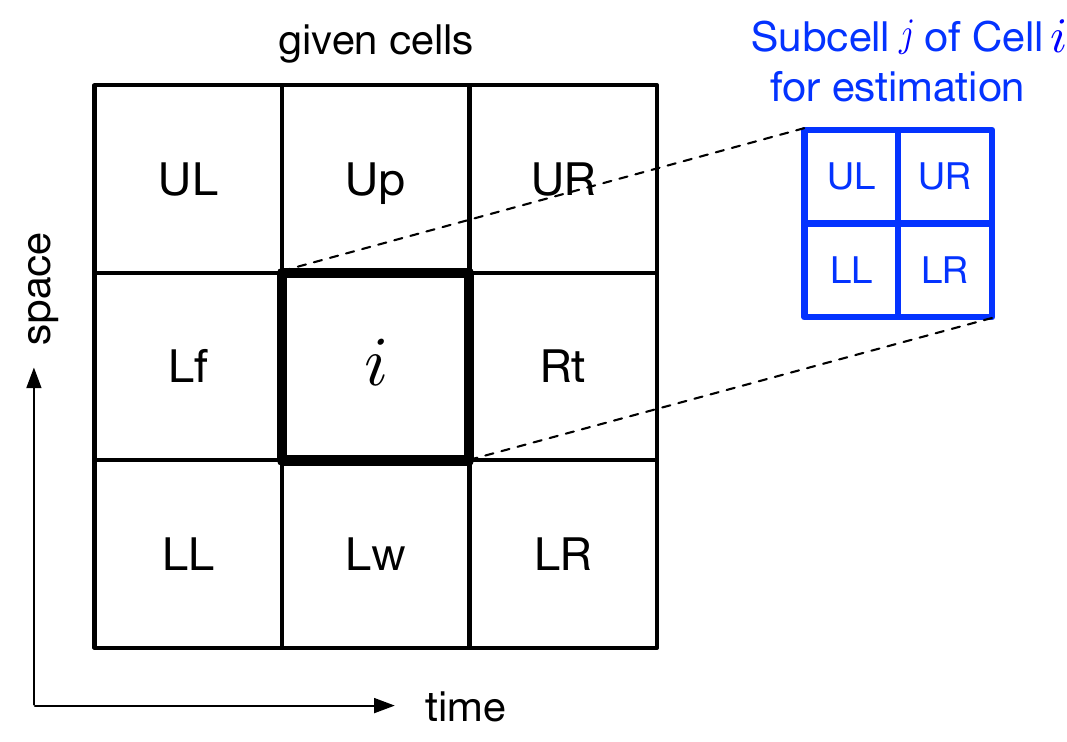}
    \caption{Schematic of the refinement of a TS diagram. 
    $i$: the cell to be refined; $j$: the subcells of Cell $i$; LL: lower-left; Lw: lower; LR: lower-right; Rt: right; UR: upper-right; Up: upper; UL: upper-left; Lf: left.}
    \label{fig:Cell}
\end{figure}

{\it Refinement problem of a TS diagram}: Given a TS diagram, partition each cell into four homogenous subcells and estimate the traffic speed of the four subcells to achieve the goal of refining the TS diagram by increasing its resolution (Figure~\ref{fig:Cell}).

\begin{figure}[h]
    \centering
    \includegraphics[width=2.1in]{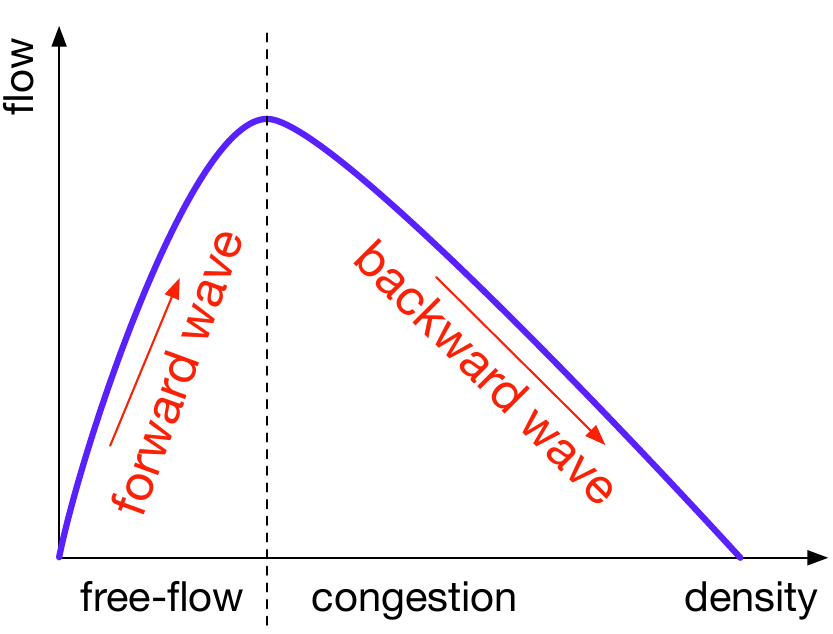}
    \caption{Fundamental diagram of traffic and traffic waves.}
    \label{fig:Wave}
\end{figure}

It is well known that there are two traffic conditions with different propagation directions of traffic waves, i.e., the free-flow condition where traffic waves propagate forward and the congested condition where traffic waves propagate backward (Figure~\ref{fig:Wave}). 
The opposite propagation directions of traffic waves have a distinguishing impact on estimating traffic states; thus, traffic conditions should be taken into account to properly capture the fundamental characteristics of traffic flow.

Denote by $i$ and $x_i$ a TS cell and its corresponding traffic speed, respectively.
Eight cells surround Cell $i$, i.e., its lower-left (LL), lower (Lw), lower-right (LR), right (Rt), upper-right (UR), upper (Up), upper-low (UL) and left (Lf) cells. 
Partition Cell $i$ into four homogenous TS subcells and denote by $y^j_i$ and $\hat y^j_{ik}$ the ground-truth and estimated speeds of Subcell $j$ of Cell $i$, where $j\in \textbf{J} = \{\text{LL}, \text{LR}, \text{UR}, \text{UL}\}$ represents the four subcells of Cell $i$ (Figure~\ref{fig:Cell}),
and $k\in \{\text{ff}, \text{cg}\}$ indicates the free-flow (ff) and congested (cg) conditions of Cell $i$.

Furthermore, involve Cell $i$ and its eight surrounding cells to propose a multiple linear regression-based refinement model and calculate $\hat y^j_{ik}$ as follows.
\begin{equation}\label{equ:y}
	\hat y^j_{ik}= \mathbf{p}^j_k \cdot\mathbf{x}_i+ \varepsilon^j_k
\end{equation}
where $\mathbf{x}_i$ is the ground-truth traffic speeds of Cell $i$ and its surrounding cells, which is written as
\begin{equation}
	\mathbf{x}_i= [x_i, x_\text{LL}, x_\text{Lw}, x_\text{LR}, x_\text{Rt}, x_\text{UR}, x_\text{Up}, x_\text{UL}, x_\text{Lf}];
\end{equation}
$\mathbf{p}^j_k$ is the regression coefficient of $\mathbf{x}_i$ and it is written as
\begin{equation}
	\mathbf{p}^j_k= [p^j_k, p^{j}_{k\text{LL}}, p^{j}_{k\text{Lw}}, p^{j}_{k\text{LR}}, p^{j}_{k\text{Rt}}, p^{j}_{k\text{UR}}, p^{j}_{k\text{Up}}, p^{j}_{k\text{UL}}, p^{j}_{k\text{Lf}}];
\end{equation}
$\varepsilon^j_k$ is a random residual term.

Given a TS diagram, the traffic condition (i.e., $k$=ff or $k$=cg) of Cell $i$ is first determined.
Then, all cells in the free-flow condition and their surrounding cells (i.e., $\mathbf{x}_i$) are used to fit Equation \ref{equ:y} with $k$=ff, and all cells in the congested condition to fit Equation 1 with $k$=cg.

According to the basic knowledge of traffic flow theory and also as selected in \cite{Treiber2002,Treiber2011a}, the speed threshold $x^*$= 60 km/h is employed to distinguish the free-flow and congested traffic conditions of Cell $i$, 
i.e., the traffic is in the free-flow condition when $ x_i>x^*$; otherwise, the traffic is in the congested condition. 
The sensitivity of the value will be further discussed in Section \ref{sec:Sensitivity}.

To obtain the ground-truth values $\mathbf{x}_i$ and $y^j_{ik}$, we employ vehicle trajectory data on a freeway segment and construct TS diagrams with various cell sizes.
To derive the value of $\mathbf{p}^j_k$, we fit Equation~\ref{equ:y} to the data using the least squares method.

\begin{figure*}
    \centering
    \includegraphics[width=7in]{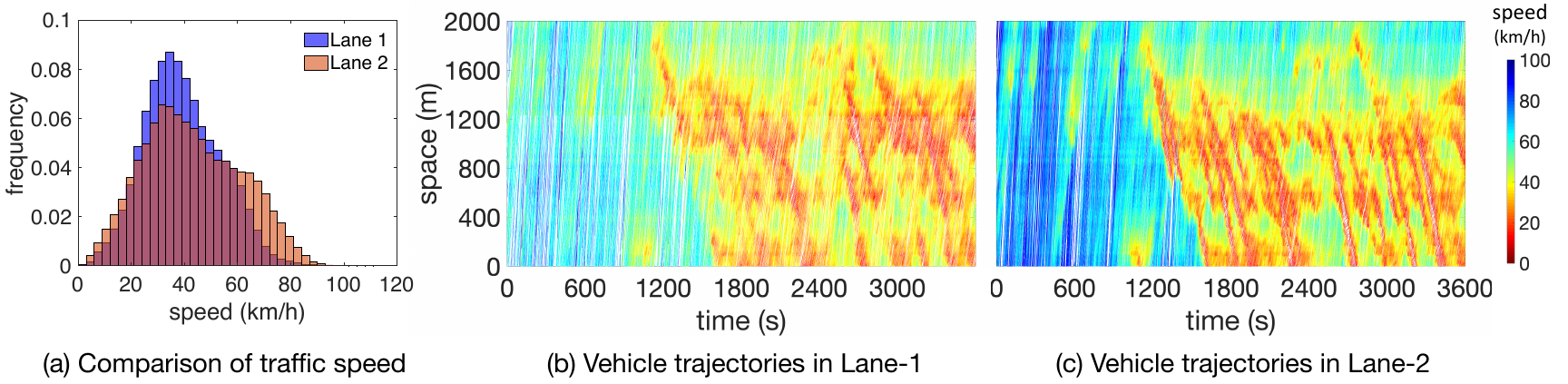}
    \caption{Summary of the 1-hour trajectory data (F001) for Ikeda-Route-11 of the Japanese ZenTraffic dataset.}
    \label{fig:Zen11}
\end{figure*}

This paper employs 1-hour 2-lane trajectory data (i.e., F001) contained in the Ikeda-Route-11 dataset of the open-source ZenTraffic dataset \cite{HanshinExp2018,Dahiyal2020} to estimate the model parameters in Equation~\ref{equ:y}.
The data was collected in 2017 on a 2-lane freeway segment of Ikeda Route 11 in Japan.
The length of the segment is 2 km, and the total duration of data collection is 5 hours (F001-F005).
The first hour of data (i.e., F001) is used to estimate the model parameters in this section, and the remaining 4 hours of data (i.e., F002-F005) are used to validate the model in Section \ref{sec:Evaluation}.
As shown in Figure~\ref{fig:Zen11}, the 1-hour data contain both free-flow traffic and congested traffic with typical traffic characteristics, such as stop-and-go waves.
It is also observed that the maximum speed in Lane-2 (approximately 90 km/h) is slightly higher than that in Lane-1 (approximately 70 km/h).
Generally speaking, the data is qualified for model fitting.

The estimated parameters with respect to different cell sizes are presented in Table~\ref{tab:Result}.
It shows that 
all $R^2$, i.e., coefficient of determination, are greater than 0.75.
In particular, $R^2$ for estimating the parameters in the congested condition are greater than 0.9.
This is quite a satisfactory result for an empirical study.

Note that, due to the limitation of space, we only take the cell sizes of 30s$\times$50m, 60s$\times$100m, 120s$\times$200m, 240s$\times$400m, 30s$\times$200m, and 60s$\times$400m as examples to present results and further tests.
In particular, the cell size of 60s$\times$400m is to align with the typical resolution of existing stationary detectors, which are usually installed several hundred meters and upload the aggregated data every 60 sec. 
All fittings with those cell sizes achieve high goodness, indicating the proposed model might not be sensitive to the cell size.
One could fit the model by following the abovementioned procedure if the needed initial cell size for refinement is not listed in Table~\ref{tab:Result}.

\begin{table*}[!htbp]\centering\footnotesize
\caption{Parameters of the refinement model estimated using the data: F001/Lanes-1 and -2/Ikeda-Route-11, ZenTraffic dataset, Japan.}\label{tab:Result}
\setlength\tabcolsep{6.7pt}
\begin{tabular}{ccccccccccccccccc}
\toprule
Cell & \multicolumn{4}{c}{Subcell} & & \multicolumn{10}{c}{Regression coefficients of Equation~\ref{equ:y}} & $R^2$\\
\cline{2-5} \cline{7-16}
 size & size & $k$ & $j$& \# & & $p^j_k$  & $p^{j}_{k\text{LL}}$ & $p^{j}_{k\text{Lw}}$ & $p^{j}_{k\text{LR}}$ & $p^{j}_{k\text{Rt}}$ & $p^{j}_{k\text{UR}}$ & $p^{j}_{k\text{Up}}$ & $p^{j}_{k\text{UL}}$ & $p^{j}_{k\text{Lf}}$ & $\varepsilon^j_k$ &  \\
\midrule
30s        & 15s       & ff     & LL & 2305 & & 0.95 & 0.43 & -0.28 & 0.02 & 0.13 & -0.27 & 0.31 & -0.11 & -0.20 & 0.84 & {\bf 0.8170} \\
$\times$   & $\times$  &        & LR & 2302 & & 0.88 & -0.38 & 0.66 & -0.12 & -0.03 & 0.25 & -0.54 & 0.06 & 0.22 & 0.30 & {\bf 0.8435} \\
50m        & 25m       &        & UR & 2303 & & 1.06 & -0.40 & 0.35 & -0.01 & -0.26 & 0.34 & -0.32 & 0.08 & 0.16 & -0.13 & {\bf 0.8489} \\
           &           &        & UL & 2306 & & 0.83 & 0.30 & -0.51 & 0.03 & 0.16 & -0.25 & 0.60 & -0.19 & 0.04 & 0.66 & {\bf 0.8171}\vspace{1mm} \\ 
           &           & cg     & LL & 7257 & & 0.92 & -0.00 & 0.21 & -0.06 & -0.19 & 0.09 & -0.20 & 0.01 & 0.20 & 0.43 & {\bf 0.9713} \\ 
           &           &        & LR & 7258 & & 0.95 & 0.03 & 0.14 & 0.05 & 0.19 & -0.06 & -0.08 & -0.01 & -0.22 & 0.41 & {\bf 0.9710} \\ 
           &           &        & UR & 7258 & & 0.99 & 0.07 & -0.20 & -0.00 & 0.21 & -0.02 & 0.20 & -0.10 & -0.16 & 0.06 & {\bf 0.9704} \\ 
           &           &        & UL & 7257 & & 0.97 & -0.08 & -0.09 & -0.01 & -0.17 & 0.00 & 0.09 & 0.05 & 0.23 & 0.19 & {\bf 0.9716} \\ 
\midrule
60s        & 30s       & ff     & LL & 532  & & 1.13 & 0.41 & -0.28 & 0.02 & 0.01 & -0.15 & 0.14 & -0.06 & -0.21 & -0.75 & {\bf 0.8585} \\
$\times$   & $\times$  &        & LR & 532  & & 0.62 & -0.39 & 0.69 & -0.07 & 0.08 & 0.11 & -0.34 & -0.01 & 0.27 & 3.65 & {\bf 0.8587} \\
100m       & 50m       &        & UR & 532  & & 0.86 & -0.41 & 0.36 & -0.05 & -0.02 & 0.15 & -0.15 & 0.04 & 0.19 & 2.38 & {\bf 0.8636} \\
           &           &        & UL & 532  & & 1.13 & 0.39 & -0.63 & 0.09 & -0.05 & -0.10 & 0.41 & -0.10 & -0.12 & -1.50 & {\bf 0.8671}\vspace{1mm} \\ 
           &           & cg     & LL & 1764 & & 0.87 & -0.05 & 0.33 & -0.12 & -0.08 & 0.07 & -0.27 & 0.08 & 0.16 & 0.52 & {\bf 0.9491} \\ 
           &           &        & LR & 1764 & & 1.04 & 0.05 & 0.02 & 0.16 & 0.05 & -0.03 & -0.04 & -0.09 & -0.14 & -0.03 & {\bf 0.9520} \\ 
           &           &        & UR & 1764 & & 0.89 & 0.06 & -0.24 & 0.05 & 0.17 & -0.06 & 0.32 & -0.17 & -0.03 & 0.86 & {\bf 0.9517} \\ 
           &           &        & UL & 1764 & & 1.02 & -0.06 & -0.03 & -0.09 & -0.09 & 0.01 & -0.01 & 0.14 & 0.10 & 0.90 & {\bf 0.9502} \\ 
\midrule
120s       & 60s       & ff     & LL & 118  & & 1.08 & 0.14 & -0.05 & 0.01 & -0.15 & -0.03 & 0.07 & -0.08 & -0.07 & 5.49 & {\bf 0.7952} \\
$\times$   & $\times$  &        & LR & 118  & & 0.58 & -0.20 & 0.53 & -0.08 & 0.18 & 0.08 & -0.30 & 0.07 & 0.18 & -1.18 & {\bf 0.7963} \\
200m       & 100m      &        & UR & 118  & & 0.75 & -0.07 & 0.15 & -0.07 & 0.19 & 0.04 & -0.01 & 0.12 & -0.06 & -1.96 & {\bf 0.7616} \\
           &           &        & UL & 118  & & 1.26 & 0.11 & -0.38 & 0.12 & -0.22 & -0.05 & 0.21 & -0.02 & -0.06 & 1.87 & {\bf 0.7812}\vspace{1mm} \\ 
           &           & cg     & LL & 404 & & 0.83 & -0.08 & 0.40 & -0.19 & -0.01 & 0.03 & -0.27 & 0.10 & 0.17 & 0.88 & {\bf 0.9246} \\ 
           &           &        & LR & 404 & & 1.07 & 0.05 & 0.02 & 0.23 & -0.08 & 0.10 & -0.10 & -0.15 & -0.13 & 0.26 & {\bf 0.9293} \\ 
           &           &        & UR & 404 & & 0.93 & 0.07 & -0.31 & 0.14 & 0.10 & -0.03 & 0.31 & -0.16 & -0.04 & 0.57 & {\bf 0.9261} \\ 
           &           &        & UL & 404 & & 1.04 & -0.03 & -0.06 & -0.13 & -0.04 & -0.08 & 0.00 & 0.21 & 0.06 & 0.97 & {\bf 0.9277} \\ 
\midrule
240s       & 120s      & ff     & LL & 19  & & 0.27 & -0.05 & 0.21 & 0.21 & -0.06 & -0.01 & 0.45 & -0.38 & 0.38 & 0.14 & {\bf 0.8752} \\
$\times$   & $\times$  &        & LR & 19  & & -0.46 & -0.49 & 1.46 & -0.46 & 0.51 & -0.76 & 0.38 & 0.29 & 0.50 & -2.13 & {\bf 0.8339} \\
400m       & 200m      &        & UR & 19  & & 0.71 & -0.40 & 0.33 & -0.28 & 0.24 & -0.18 & 0.24 & 0.36 & 0.05 & -3.23 & {\bf 0.8273} \\
           &           &        & UL & 19  & & 2.54 & 0.48 & -1.38 & 0.43 & -0.51 & 0.70 & -0.61 & -0.26 & -0.39 & 1.88 & {\bf 0.8602}\vspace{1mm} \\
           &           & cg     & LL & 85  & & 0.76 & -0.01 & 0.47 & -0.22 & 0.08 & -0.10 & -0.03 & 0.09 & -0.03 & -0.49 & {\bf 0.9353} \\ 
           &           &        & LR & 85  & & 1.37 & 0.10 & -0.29 & 0.29 & -0.23 & 0.06 & -0.30 & 0.08 & -0.18 & 4.32 & {\bf 0.9266} \\ 
           &           &        & UR & 85  & & 0.82 & -0.03 & -0.26 & 0.02 & 0.10 & -0.04 & 0.44 & -0.22 & 0.16 & -0.76 & {\bf 0.9514} \\ 
           &           &        & UL & 85  & & 0.97 & -0.06 & 0.13 & -0.11 & 0.03 & 0.04 & -0.02 & 0.03 & 0.09 & -1.88 & {\bf 0.9344} \\ 
\midrule
30s        & 15s       & ff     & LL & 512 & & 1.10 & 0.33 & -0.20 & 0.02 & 0.02 & -0.12 & 0.02 & 0.01 & -0.19 & 1.09 & {\bf 0.8881} \\ 
$\times$   & $\times$  &        & LR & 512 & & 0.65 & -0.26 & 0.62 & -0.17 & 0.08 & 0.15 & -0.27 & 0.02 & 0.14 & 3.36 & {\bf 0.8849} \\ 
200m       & 100m      &        & UR & 511 & & 1.26 & -0.23 & 0.09 & 0.01 & -0.19 & 0.19 & -0.16 & 0.05 & -0.01 & -0.38 & {\bf 0.8930} \\ 
           &           &        & UL & 512 & & 0.71 & 0.25 & -0.32 & 0.01 & 0.16 & -0.23 & 0.45 & -0.15 & 0.10 & 1.05 & {\bf 0.8638} \vspace{1mm} \\
           &           & cg     & LL & 1696 & & 0.86 & -0.09 & 0.25 & 0.01 & -0.22 & 0.08 & -0.06 & -0.17 & 0.32 & 0.89 & {\bf 0.9480} \\ 
           &           &        & LR & 1696 & & 1.08 & -0.03 & 0.12 & 0.13 & 0.06 & 0.08 & -0.26 & 0.02 & -0.19 & 0.62 & {\bf 0.9523} \\ 
           &           &        & UR & 1696 & & 0.93 & 0.09 & -0.11 & -0.14 & 0.32 & -0.09 & 0.19 & 0.02 & -0.22 & 0.59 & {\bf 0.9541} \\ 
           &           &        & UL & 1696 & & 0.99 & 0.04 & -0.25 & 0.03 & -0.13 & -0.03 & 0.06 & 0.11 & 0.17 & 0.77 & {\bf 0.9561} \\ 
\midrule
60s        & 30s       & ff     & LL & 98  & & 1.16 & 0.38 & -0.19 & 0.01 & -0.03 & -0.10 & -0.03 & 0.05 & -0.28 & 0.86 & {\bf 0.8976} \\
$\times$   & $\times$  &        & LR & 98  & & 0.67 & -0.26 & 0.53 & -0.07 & 0.02 & 0.11 & -0.16 & 0.06 & -0.04 & 10.20 & {\bf 0.8525}  \\
400m       & 200m      &        & UR & 98  & & 1.24 & -0.22 & 0.01 & 0.02 & -0.11 & 0.19 & -0.07 & 0.05 & -0.10 & -0.01 & {\bf 0.9063} \\
           &           &        & UL & 98  & & 0.76 & 0.19 & -0.35 & 0.04 & 0.18 & -0.20 & 0.30 & -0.17 & 0.30 & -3.01 &  {\bf 0.8215} \vspace{1mm} \\
           &           & cg     & LL & 386 & & 0.77 & -0.08 & 0.07 & 0.13 & -0.20 & 0.09 & -0.14 & -0.13 & 0.42 & 3.78 & {\bf 0.9140} \\
           &           &        & LR & 386 & & 1.18 & -0.01 & -0.02 & 0.17 & 0.03 & 0.06 & -0.34 & 0.10 & -0.22 & 2.33 & {\bf 0.9300} \\
           &           &        & UR & 386 & & 0.86 & 0.05 & 0.03 & -0.22 & 0.29 & 0.04 & 0.10 & 0.02 & -0.14 & -0.55 & {\bf 0.9204} \\
           &           &        & UL & 386 & & 0.99 & 0.11 & -0.20 & -0.02 & -0.08 & -0.13 & 0.33 & 0.00 & 0.06 & -2.43 & {\bf 0.9456} \\           
\bottomrule
\end{tabular}
\end{table*}

\section{Model evaluation}\label{sec:Evaluation}

\subsection{Data for model tests}\label{sec:Data}

To fairly test the model (Equation~\ref{equ:y} and Table~\ref{tab:Result}) and avoid the autocorrelation of data, we employ the following data collected from completely different freeway segments for model tests.
It is believed that the three datasets could well guarantee the reliability of the evaluation and the transferability of the proposed model, which will be discussed in more detail in Section \ref{sec:Transferability}.

\begin{itemize}    
    \item 
    {\it Different-time data}. The other 4-hour 2-lane vehicle trajectory data (i.e., F002-F005) in the Ikeda-Route-11 dataset of the ZenTraffic dataset in Japan \cite{HanshinExp2018,Dahiyal2020}.
    
    \vspace{1mm}
    
    \item 
    {\it Different-location data}. The 5-hour 2-lane vehicle trajectory data (i.e., F001-F005) in the Wangan-Route-4 dataset of the ZenTraffic dataset in Japan \cite{HanshinExp2018,Dahiyal2020}.
    The Wangan-Route-4 dataset is another high-fidelity trajectory dataset contained in the ZenTraffic dataset. 
    The data was collected on a 1.6 km 2-lane freeway segment in Wangan Route 4 in Japan, and the duration of data collection was 5 hours in total.
    
    \vspace{1mm}
    
    \item 
    {\it Different-country data}. The 45-minute 5-lane trajectory vehicle data of the US-101 dataset was provided by the Next Generation Simulation (NGSIM) program \cite{NGSIM2006}. 
    The well-known US-101 dataset was collected on a 640 m 5-lane segment in the vicinity of Lankershim Avenue on the southbound US-101 freeway in Los Angeles, California, from 7:50 a.m. to 8:35 a.m. on June 15, 2005.
        
\end{itemize}

\begin{table*}[!htbp]\centering\footnotesize
\caption{Attributes of the trajectory datasets utilized for model estimation and evaluation.}\label{tab:Data}
\setlength\tabcolsep{17pt}
\begin{tabular}{cccccc}
\toprule
 & Dataset & Segment length (km) & Collection duration (h) & Lane \# & Country  \vspace{1mm}\\
\midrule
Model fitting & Ikeda-Route-11 & 2 & 1 (F001) & 2 & Japan\vspace{1mm}\\
\midrule
Evaluation & Ikeda-Route-11 & 2 & 4 (F002-F005) & 2 & Japan\vspace{1mm}\\
& Wangan-Route-4 & 1.6 & 5 (F001-F005) &  2 & Japan\vspace{1mm}\\
& US-101 & 0.64 & 0.75 (7:50-8:35) & 5 & United States\vspace{1mm}\\
& \textbf{Total} & \textbf{4.24} & \textbf{9.75} & \textbf{9} \\
\bottomrule
\end{tabular}
\end{table*}

\subsection{Measures of errors}\label{sec:Measures}

\begin{table*}[!htbp]\centering\footnotesize
\caption{Results of model evaluation: 1-4 test.}\label{tab:Result14}
\setlength\tabcolsep{8pt}
\begin{tabular}{ccccccccccccccc}
\toprule
Cell & \multicolumn{2}{c}{1-4 subcell} & & \multicolumn{3}{c}{Ikeda-Route-11 (F002-F005)} & & \multicolumn{3}{c}{Wangan-Route-4 (F001-F005)} & & \multicolumn{3}{c}{US-101} \\
\cline{2-3} \cline{5-7} \cline{9-11}\cline{13-15}
 size & size & $j$ &  & Subcell \#  & MAE  & MAPE  & & Subcell \#  & MAE  & MAPE & & Subcell \#  & MAE  & MAPE\\
\midrule
30s      & 15s      & LL   &&  38028 & 1.915 & {\bf 0.040}   &&  36941 & 1.603 & {\bf 0.032}   &&   4894 & 2.158 & {\bf 0.086}   \\ 
$\times$ & $\times$ & LR   &&  38035 & 1.911 & {\bf 0.040}   &&   36948 & 1.601 & {\bf 0.032}   &&   4894 & 2.157 & {\bf 0.085}   \\ 
50m      & 25m      & UR   &&  38025 & 1.904 & {\bf 0.040}   &&  36946 & 1.609 & {\bf 0.032}   &&   4894 & 2.144 & {\bf 0.080}   \\ 
         &          & UL   &&  38025 & 1.919 & {\bf 0.041}   &&  36947 & 1.611 & {\bf 0.032}   &&   4894 & 2.211 & {\bf 0.085}   \\ 
\midrule
60s      & 30s      & LL   &&  9153 & 2.292 & {\bf 0.050}   &&   8790 & 1.911 & {\bf 0.039}    &&   1100 & 2.823 & {\bf 0.098}   \\ 
$\times$ & $\times$ & LR   &&   9153 & 2.267 & {\bf 0.050}   &&   8790 & 1.929 & {\bf 0.040}   &&   1100 & 2.748 & {\bf 0.093}   \\ 
100m     & 50m      & UR   &&   9153 & 2.229 & {\bf 0.049}   &&   8790 & 1.925 & {\bf 0.039}   &&   1100 & 2.754 & {\bf 0.095}   \\ 
         &          & UL   &&   9153 & 2.301 & {\bf 0.049}   &&   8790 & 1.947 & {\bf 0.039}   &&   1100 & 2.927 & {\bf 0.099}   \\ 
\midrule              
120s     & 60s      & LL   &&   2088 & 2.783 & {\bf 0.061}   &&  1974 & 2.228 & {\bf 0.046}   && ---  & ---  & ---   \\ 
$\times$ & $\times$ & LR   &&   2088 & 2.781 & {\bf 0.063}   &&   1974 & 2.242 & {\bf 0.047}  && ---  & --- & ---   \\ 
200m     & 100m     & UR   &&   2088 & 2.660 & {\bf 0.061}   &&  1974 & 2.243 & {\bf 0.044}   && ---  & --- & ---  \\ 
         &          & UL   &&   2088 & 2.782 & {\bf 0.061}  &&   1974 & 2.232 & {\bf 0.044}   && ---  & --- & ---  \\ 
\midrule              
240      & 120s     & LL   &&   416 & 3.316 & {\bf 0.073}   &&  390 & 2.916 & {\bf 0.058}   && ---  & --- & ---  \\ 
$\times$ & $\times$ & LR   &&   416 & 4.431 & {\bf 0.085}   &&  390 & 4.543 & {\bf 0.084}   && ---  & --- & ---  \\ 
400m     & 200m     & UR   &&   416 & 3.634 & {\bf 0.077}   &&  390 & 3.419 & {\bf 0.062}   && ---  & --- & ---  \\ 
         &          & UL   &&   416 & 4.434 & {\bf 0.085}   &&  390 & 4.298 & {\bf 0.079}   && ---  & --- & ---  \\  
\midrule 
30s      & 15s      & LL   &&   8789 & 2.132 & {\bf 0.049}   &&  8344 & 1.715 & {\bf 0.036}  && ---  & --- & ---  \\ 
$\times$ & $\times$ & LR   &&   8790 & 2.221 & {\bf 0.050}   &&  8344 & 1.864 & {\bf 0.038}  && ---  & --- & ---  \\ 
200m     & 100m     & UR   &&   8791 & 2.059 & {\bf 0.047}   &&  8344 & 1.781 & {\bf 0.035}  && ---  & --- & ---  \\ 
         &          & UL   &&   8789 & 2.182 & {\bf 0.048}   &&  8344 & 1.902 & {\bf 0.037}  && ---  & --- & ---  \\    
\midrule 
60s      & 30s      & LL   &&   1925 & 2.836 & {\bf 0.064}   &&  1758 & 2.260 & {\bf 0.045}  && ---  & --- & ---  \\ 
$\times$ & $\times$ & LR   &&   1925 & 2.712 & {\bf 0.060}   &&  1758 & 2.361 & {\bf 0.047}  && ---  & --- & ---  \\ 
400m     & 200m     & UR   &&   1925 & 2.678 & {\bf 0.063}   &&  1758 & 2.512 & {\bf 0.048}  && ---  & --- & ---  \\ 
         &          & UL   &&   1925 & 2.640 & {\bf 0.059}   &&  1758 & 2.418 & {\bf 0.046}  && ---  & --- & ---  \\            
\bottomrule
\end{tabular}
\begin{tablenotes}\footnotesize
\item[1]  \hspace{-3mm} Note: the length of the segment where the US-101 dataset was collected is not enough for the test with cell sizes 120s$\times$200m and 240s$\times$400m.
\end{tablenotes}   
\end{table*}

The following measures of errors are employed to reflect the accuracy of the refinement. 
\begin{itemize}    
    \item Mean Absolute Error: 
    
\begin{equation}\label{equ:mae}
	\text{MAE}=  \frac{1}{N|\textbf{J}|}\sum_{i=1}^N  \sum_{j\in\textbf{J}} \left | y^j_i - \hat y^j_{ik}  \right |
\end{equation}
where $|\textbf{J}|$ is the size of set $\textbf{J}$ and $|\textbf{J}|=4$.

\vspace{4mm}

    \item Mean Absolute Percentage Error:
\begin{equation}\label{equ:mape}
	\text{MAPE}=  \frac{1}{N|\textbf{J}|}\sum_{i=1}^N  \sum_{j\in\textbf{J}} \left | \frac{y^j_i - \hat y^j_{ik}}{y^j_i}  \right |
\end{equation}

\end{itemize}

\subsection{Evaluation: 1-4 test}

To validate the proposed refinement model, we first conduct a ``1-4 test" as follows:
Given a TS cell, we estimate the traffic speed of its 4 homogenous subcells so that the resolution of the corresponding TS diagram is increased 4 times.

Table~\ref{tab:Result14} presents the results.
All MAPEs are less than 0.1, even for the data collected from another country (i.e., US-101 dataset), and many of the MAPEs are even less than 0.05.
The results demonstrate that the proposed model achieves high refinement accuracy in the 1-4 test.
In addition, it is observed that the refinement errors increase with the enlargement of the cells of the TS diagram to be refined.
The reason is easy to understand: The larger the initial TS cells are, the more traffic information is lost in the construction of the initial TS diagram, and thus, it is more difficult to obtain an accurate refinement.

\subsection{Evaluation: 1-4-16 test}

Furthermore, we conduct a stricter ``1-4-16 test" as follows: 
Given a TS cell, we estimate the traffic speed of its homogenous 4 subcells (referred to as 1-4 subcells) and further estimate the traffic speed of the homogenous 4 subcells (referred to as 4-16 subcells) of each of the 1-4 subcells by using the estimated speed of these 1-4 subcells.
In the test, the resolution of a given TS diagram will be increased 16 times.
Note that we only input the estimated speed of 1-4 subcells when estimating the traffic speed of 4-16 subcells, meaning that, in the 1-4-16 test, the input data only consists of the speeds of the TS cells in the initial TS diagram.

Table~\ref{tab:Result146} presents the results, from which the following observations can be made: 
All MAPEs are less than 0.17, and many MAPEs are less than 0.1.
Compared with the 1-4 test, increasing the resolution 16 times indeed results in more errors, whereas the magnitude of the errors is acceptable.  
In general, the estimation results are satisfactory, in particular for a study that is based on empirical data.

\begin{table*}[!htbp]\centering\footnotesize
\caption{Results of model evaluation: 1-4-16 test.}\label{tab:Result146}
\setlength\tabcolsep{7.8pt}
\begin{tabular}{ccccccccccccccc}
\toprule
Cell & \multicolumn{2}{c}{1-4-16 subcell} & & \multicolumn{3}{c}{Ikeda-Route-11 (F002-F005)} & & \multicolumn{3}{c}{Wangan-Route-4 (F001-F005)} & & \multicolumn{3}{c}{US-101} \\
\cline{2-3} \cline{5-7} \cline{9-11}\cline{13-15}
 size & size & $j$ &  & Subcell \#  & MAE  & MAPE  & & Subcell \#  & MAE  & MAPE & & Subcell \#  & MAE  & MAPE\\
\midrule
60s      & 15s      & LL &&   31994 & 3.195 & {\bf 0.070}  &&   29893 & 2.606 & {\bf0.054}   &&   3006 & 3.804 & {\bf0.160}   \\ 
$\times$ & $\times$ & LR &&   31987 & 3.218 & {\bf0.073}   &&   29899 & 2.658 & {\bf0.055}  &&   3006 & 3.787 & {\bf0.158}   \\ 
100m     & 25m      & UR &&   31987 & 3.218 & {\bf0.073}   &&   29897 & 2.605 & {\bf0.053}  &&   3006 & 3.783 & {\bf0.153}   \\ 
         &          & UL &&   31990 & 3.240 & {\bf0.072}   &&   29897 & 2.611 & {\bf0.053}   &&   3006 & 3.894 & {\bf0.155}   \\ 
\midrule              
120s     & 30s      & LL &&   6272 & 3.953 & {\bf0.093}   &&  5440 & 3.130 & {\bf0.065}   && ---  & ---  & ---   \\ 
$\times$ & $\times$ & LR &&   6272 & 3.941 & {\bf0.097}   &&  5440 & 3.110 & {\bf0.066}   && ---  & --- & ---   \\ 
200m     & 50m      & UR &&   6272 & 3.813 & {\bf0.091}   &&  5440 & 2.923 & {\bf0.061}   && ---  & --- & ---  \\ 
         &          & UL &&   6272 & 3.894 & {\bf0.090}   &&  5440 & 3.058 & {\bf0.063}   && ---  & --- & ---  \\ 
\midrule              
240s     & 60s      & LL &&   768 & 5.808 & {\bf0.132}   &&  480 & 4.002 & {\bf0.085}   && ---  & --- & ---  \\ 
$\times$ & $\times$ & LR &&   768 & 7.720 & {\bf0.160}   &&  480 & 6.134 & {\bf0.122}  && ---  & --- & ---  \\ 
400m     & 100m     & UR &&   768 & 6.254 & {\bf0.137}   &&  480 & 4.776 & {\bf0.098}  && ---  & --- & ---  \\ 
         &          & UL &&   768 & 7.988 & {\bf0.166}   &&  480 & 6.093 & {\bf0.120}   && ---  & --- & ---  \\                                                        
\midrule              
60s      & 15s      & LL &&   3732 & 4.401 & {\bf 0.109}   &&  2300 & 3.300 & {\bf 0.071}   && ---  & --- & ---  \\ 
$\times$ & $\times$ & LR &&   3735 & 4.373 & {\bf 0.109}   &&  2300 & 3.210 & {\bf 0.070}  && ---  & --- & ---  \\ 
400m     & 100m     & UR &&   3735 & 4.371 & {\bf 0.109}   &&  2300 & 3.225 & {\bf 0.071}  && ---  & --- & ---  \\ 
         &          & UL &&   3734 & 4.251 & {\bf 0.105}   &&  2300 & 3.155 & {\bf 0.070}   && ---  & --- & ---  \\                                                                 
\bottomrule
\end{tabular}
\begin{tablenotes}\footnotesize
\item[1]  \hspace{-2mm} Note: the length of the segment where the US-101 dataset was collected is not enough for the test with cell sizes 120s$\times$200m and 240s$\times$400m.
\end{tablenotes}   
\end{table*}

\begin{figure*}
    \centering
    \includegraphics[width=7in]{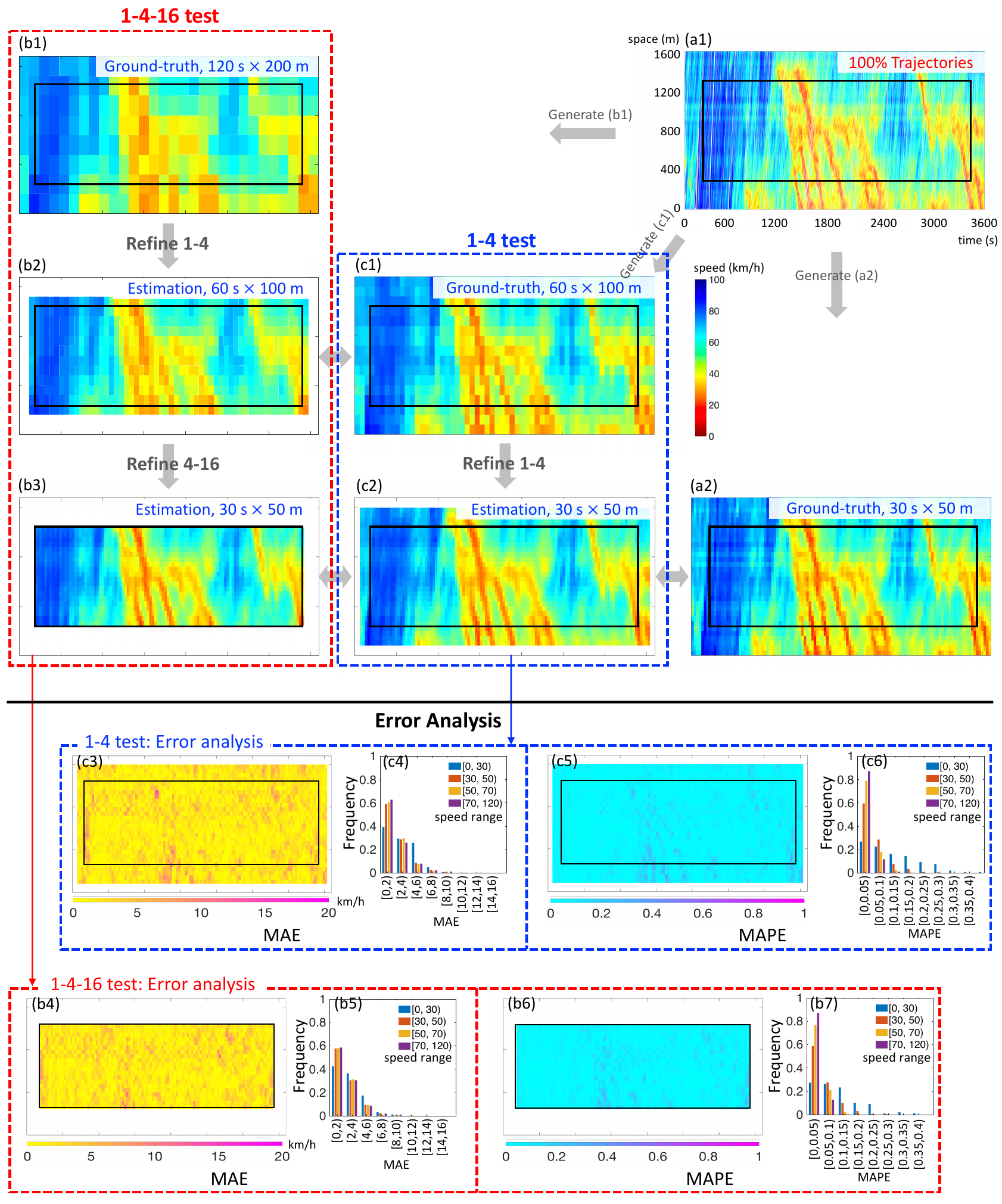}
    \caption{Visual presentation of the 1-4 test (red dashed box) and the 1-4-16 test (blue dashed box) with a selected example.
    {\bf Target data} (a1)(a2): F003/Lane-2/Wangan-Route-4.
    {\bf 1-4-16 test} (b1)(b2)(b3): Refining a TS diagram from 120s$\times$200m cells to 30s$\times$50m cells, i.e., increasing the resolution 16 times.
    {\bf 1-4 test} (c1)(c2): Refining a TS diagram from 60s$\times$100m cells to 30s$\times$50m, i.e., increasing the resolution 4 times. 
    {\bf Error analysis of 1-4-16 test}: (b4) TS diagram of MAE; (b5) Frequency of MAE in different ranges of cell speed; (b6) TS diagram of MAPE; (b7) Frequency of MAPE in different ranges of cell speed.
    {\bf Error analysis of 1-4 test}: (c3) TS diagram of MAE; (c4) Frequency of MAE in different ranges of cell speed; (c5) TS diagram of MAPE; (c6) Frequency of MAPE in different ranges of cell speed.
    Note that the edge of a TS diagram cannot be refined since the surrounding cells have to be taken as the input data. 
    Therefore, the resulting TS diagram is smaller in size than the initial diagram.
    For convenience in our comparison, we add a black box to each diagram to indicate the resulting region in the 1-4-16 test (b3).}    
    \label{fig:STPlot146}
\end{figure*}

To visually present the results, we illustrate an example of the 1-4 and 1-4-16 tests in Figure~\ref{fig:STPlot146}.
It is impressive to observe that a TS diagram with quite coarse traffic patterns (Figure~\ref{fig:STPlot146}(b1)) can be refined to a TS diagram where even detailed stop-and-go waves can be observed (Figure~\ref{fig:STPlot146}(b3)).
Moreover, in the low-resolution TS diagram (Figure~\ref{fig:STPlot146}(b1)), the traffic propagation directions are almost vertical. 
After the refinement, the propagation directions with slopes are clearly revealed in Figures~\ref{fig:STPlot146}(b3) and \ref{fig:STPlot146}(c2), which is more consistent with the real-world traffic patterns (Figures~\ref{fig:STPlot146}(a) and \ref{fig:STPlot146}(d)). 
Compared with the ground-truth diagram (Figure~\ref{fig:STPlot146}(d)), 
both the refined TS diagrams in the 1-4 and 1-4-16 tests (Figures~\ref{fig:STPlot146}(c2) and \ref{fig:STPlot146}(b3)) satisfactorily reproduce the details of traffic patterns such as stop-and-go waves and wave directions.

To better understand the refinement errors, we visualize the errors in the lower part of Figure~\ref{fig:STPlot146}.
From the TS diagrams of the estimation errors in Figures~\ref{fig:STPlot146}(b4)(b6)(c3)(c5), it can be seen that the errors homogeneously distribute in the TS plane, indicating that the proposed model will not result in significantly large bias in the refinement of some specific traffic states, including the boundary between the free-flow and congested conditions.  
Similarly, the histograms in Figures~\ref{fig:STPlot146}(b5)(b7)(c4)(c6) further confirm that the magnitude of the estimation errors has no clear difference when refining the TS cells with various traffic states.
Note that the relatively large MAPEs for lower traffic speed are mainly resulted from the smaller denominator (i.e., the ground-truth traffic speed) in Equation \ref{equ:mape}.
In general, the visual comparison clearly demonstrates the power of the proposed refinement model.

Table~\ref{tab:Result146} shows that the MAPEs for refining the diagram constructed using the US-101 dataset are slightly greater than those of the other diagrams.
The reasons are discussed as follows:
In comparison, the MAEs for refining the US-101 diagram are not significantly greater than those of the other diagrams, indicating that the high MAPEs partially result from the lower ground-truth traffic speed (i.e., the denominator of Equation~\ref{equ:mape}) that is contained by the US-101 dataset.
In addition, there might still be certain differences between the traffic patterns of two different countries or among the specific segments of data collection. 
The existence of errors is understandable, and the magnitude is acceptable.

\begin{figure}[!htbp]
    \centering
    \includegraphics[width=3.5in]{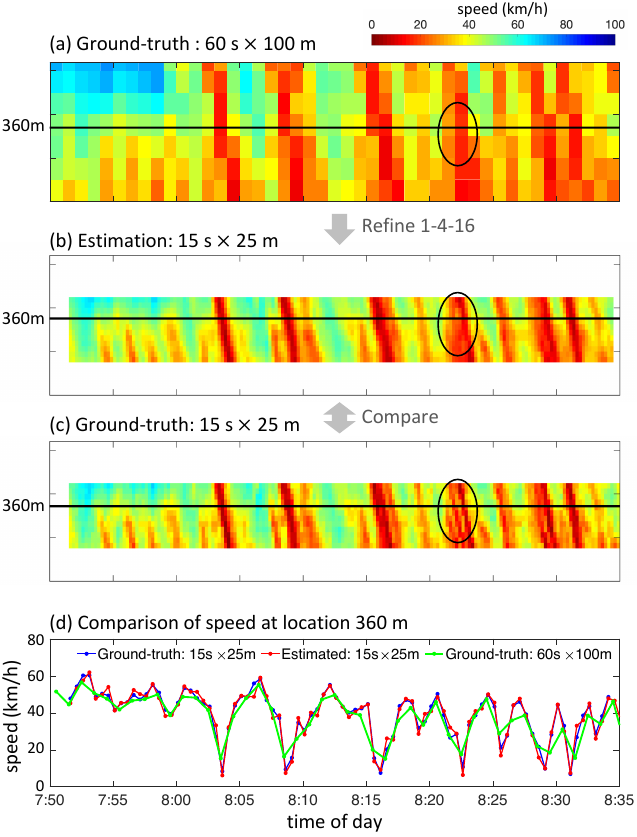}
    \caption{1-4-16 test of the 60s$\times$100m TS diagram constructed using the US-101 Lane-2 data.
    }
    \label{fig:US101}
\end{figure}

Figure~\ref{fig:US101} is supplemented to visually show the details of the refinement results of the US-101 diagram in the 1-4-16 test. 
The main traffic patterns are well retained in the 15s$\times$25m TS diagram (Figure~\ref{fig:US101}(b)) that is refined from the 60s$\times$100m TS diagram (Figure~\ref{fig:US101}(a)) compared with the ground-truth TS diagram in Figure~\ref{fig:US101}(c).
Figure~\ref{fig:US101}(d), which presents the speed changes for a specific location in the TS diagrams, further supports the judgment,
i.e., the speed changes of the refined TS diagram (red line) generally overlap those of the ground-truth diagram (blue line), and it shows a clearer up-and-down trend, i.e., stop-and-go waves, compared with that in the initial TS diagram with lower resolution (green line). 
Moreover, as indicated by the circles in Figures~\ref{fig:US101}(b) and \ref{fig:US101}(c), the refined TS diagram fails to capture sufficient detail in traffic propagation.
This lack of detail might be attributed to the notion that the small amount of detail has already been omitted when constructing the initial low-resolution TS diagram (refer to the circled area in Figure~\ref{fig:US101}(a)), making it very difficult or even impossible to be reproduced.


\section{Model comparison}\label{sec:Comparison}

It is curious to know if traditional TSE methods are competent in refining TS diagrams, although a comparative method that was proposed directly for the refinement problem might not be found at present.
To the end, we select the classic adaptive smoothing method (ASM) \cite{Treiber2002,Treiber2011a} with the same parameter values (Table \ref{tab:Parameter}) in \cite{Treiber2011a} to make a comparison.

\begin{table}[htbp]\centering\footnotesize
    \centering \caption{Parameters of the adaptive smoothing method.}\label{tab:Parameter}
    \setlength\tabcolsep{2pt}
    \begin{tabular}{lr}
    \toprule
	Parameter & Value \\
    \midrule
	smoothing width in time coordinate     &   $\sigma=60$ s\\
	smoothing width in space coordinate   &  $\tau=100$ m\\
	propagation speed of perturbations in free traffic   &  $c_\text{free}=70$ km/h\\
	propagation speed of perturbations in congested traffic   &  $c_\text{cong}=-15$ km/h\\	
	crossover from congested to free traffic  & $V_\text{thr}=60$ km/h\\
	transition width between congested and free traffic & $\Delta V= 20$ km/h\\
    \bottomrule
    \end{tabular}
\begin{tablenotes}\footnotesize
\item[1]  \hspace{-4mm}  Note: the values of $\sigma$ and $\tau$ are assigned according to the rule given in \cite{Treiber2011a}, 
\item[2]  \hspace{-4mm}  i.e., half of the spacing and sampling time of the input data 120s$\times$200m.
\item[3]  \hspace{-4mm}  Other values are the same as the ones in \cite{Treiber2011a}.
\end{tablenotes} 
\end{table}

\begin{figure}[!htbp]
    \centering
    \includegraphics[width=2.4in]{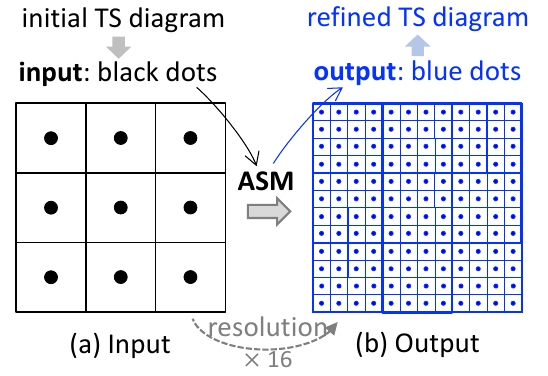}
    \caption{Schematic of applying the ASM to solve the refinement problem.}
    \label{fig:ASM_Method}
\end{figure}

The TS diagram that is constructed using the data F003/Lane-2/Wangan-Route-4 is taken as the input of the ASM.
To make it suitable for the ASM, the traffic speed of a TS cell is treated as the one located at the central dot of the TS cell, resulting in a series of speed data that scatter in the TS plane (Figure~\ref{fig:ASM_Method}(a)).
The target is to estimate the traffic speed at the homogeneously-distributed TS dots whose number is 16 times more than that of the initial dots (Figure~\ref{fig:ASM_Method}(b)), i.e., increasing the resolution 16 times.
With such input and target, the result of the ASM is comparable with the 1-4-16 test in Table \ref{tab:Result146} and Figure~\ref{fig:STPlot146}(b3).

\begin{figure}[!htbp]
    \centering
    \includegraphics[width=3in]{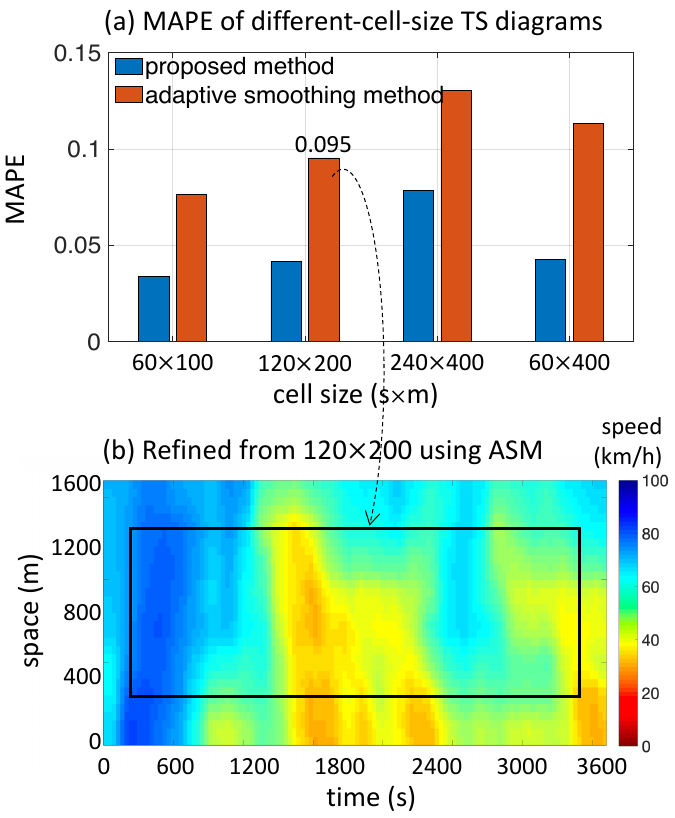}
    \caption{Results of model comparison: Proposed model vs. ASM.}
    \label{fig:ASM}
\end{figure}

Figure~\ref{fig:ASM}(a) compares the resulting MAPEs when refining the TS diagrams with the cell sizes of 60s$\times$100m, 120s$\times$200m, 240s$\times$400m, and 60s$\times$400m. 
The MAPEs resulted from the proposed refinement method are much smaller than those from the ASM. 
The MAPEs are even less than half of those resulted from the ASM when the cell sizes are 60s$\times$100m, 120s$\times$200m, and 60s$\times$400m.
It is worth noting that the proposed method still outperforms the ASM in refining the TS diagram whose resolution (i.e., 60s$\times$400m) is specifically selected to align with the typical resolution of existing stationary detectors, i.e., the main ``battlefield" of the ASM.
In addition, Figure~\ref{fig:ASM}(b) presents the 30s$\times$50m TS diagram refined from 120s$\times$200m by using the ASM.
Compared to the one resulted from the proposed method in Figure~\ref{fig:STPlot146}(b3), the one refined by the ASM does not clearly display important traffic characteristics.  
It is not difficult to understand that the ASM, which is based on a continuous filter, tends to generate smoothing results, making a TS diagram look quite ``foggy".
Therefore, based on the above comparison, we might conclude that the proposed method outperforms the ASM when refining a TS diagram.

\section{Discussion}\label{sec:Discussion}

\subsection{Regarding the sensitivity of the speed threshold of distinguishing free-flow and congested traffic conditions}\label{sec:Sensitivity}

\begin{figure}[!htbp]
    \centering
    \includegraphics[width=3in]{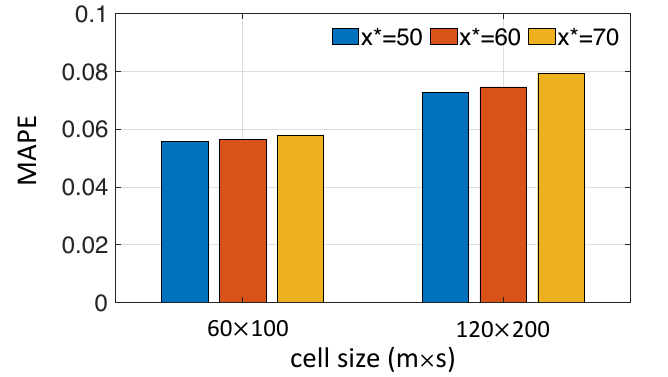}
    \caption{Error comparison of 1-4-16 tests with different initial cell sizes (i.e., 60s$\times$100m and 120s$\times$200m) and different speed thresholds (i.e., $x^*$=50, $x^*$=60, $x^*$=70) of distinguishing free-flow and congested traffic conditions. 
    The target data for the refinement is F003/Lane-2/Wangan-Route-4.}
    \label{fig:SpeedThreshold}
\end{figure}

In the proposed model, there is only one exogenous parameter, i.e., the speed threshold $x^*$ of distinguishing free-flow and congested traffic conditions.
We now test if the refinement result is sensitive to the value: 
Set $x^*$=50, $x^*$=60, $x^*$=70, respectively, and re-estimate the model parameters in Equation~\ref{equ:y} using the data F001/Lane-1 and -2/Ikeda-Route-11; it is exactly the same as the procedure and data used in Section \ref{sec:Model}.
Then, we use the re-estimated models to refine the different-cell-size TS diagrams constructed using the data F003/Lane-2/Wangan-Route-4. 
Figure~\ref{fig:SpeedThreshold} presents the results, and it can be seen that the values of $x^*$ do not significantly change the refinement accuracy, turning out that the proposed model is not sensitive to the value of its only exogenous parameter.

\subsection{Regarding the necessity of considering traffic conditions}
Distinguishing free-flow and congested traffic conditions is indispensable for the proposed refinement model.
In the preliminary stage of the study, we did not distinguish the conditions and failed to capture the forward waves in free-flow traffic (i.e., we do not observe free-flow waves toward the top right corner in, e.g., Figure~\ref{fig:STPlot146}(c2)), although a high goodness of fit could also be achieved.

\subsection{Regarding the simpleness of the method}
We choose a linear regression method, which is simple in form, as the foundation of modeling based on the following considerations:
First, as we have carefully shown, the simple method is able to achieve high goodness of fits (Table~\ref{tab:Result}) and make satisfactory refinement, even for estimating the TS diagram from another country (Table~\ref{tab:Result14}) as well as in the stricter 1-4-16 test (Table~\ref{tab:Result146} and Figure~\ref{fig:STPlot146}). 
Achieving good results is the ultimate goal for a study that aims to solve a problem instead of proposing a ``fancy" method.
Second, a linear regression method could give a distinct expression that is simply contained in Table~\ref{tab:Result}; it is very convenient for spread and applications in practice.
We could also obtain the dominant impact factors from the parameter values. 
%

\subsection{Regarding the transferability of the method}\label{sec:Transferability}
It is believed that the proposed refinement model has good transferability.
First, good transferability has been demonstrated in the tests with data collected at different times, in different locations and even in different countries.
It is even left-hand traffic in Japan, which is quite different from the traffic in the United States.
Second, diverse and abundant traffic patterns have been involved in the evaluation. 
In particular, the traffic in the Japanese datasets shows clear discontinuity resulted from merging vehicles from on-ramps and the phantom jams that occurred in the middle of the segment. One may refer to the time-space diagrams of trajectories published on their website \cite{HanshinExp2018}.
Third, the three datasets cover a total of 4.24 km segments, a 9.75 h collection duration, and a total of 9 lanes (Table~\ref{tab:Data}). 
Such a large spatiotemporal range of the data ensures the sufficiency of the evaluation.

Beyond adding new tests, good transferability could be actually guaranteed by the traffic characteristic itself.
It is well known that traffic flow shows similar and limited characteristics worldwide; for example, the backward wave speed is globally observed between -20 and -10 km/h, and the period of the stop-and-go wave ranges from 2 to 15 min \cite{Zielke2008,Chen2012i,Jiang2014,He2015a}.
Many existing works have achieved satisfactory estimations by simply inputting typical traffic characteristics \cite{Treiber2011a,Wang2005,Laval2012}, and thus there may be no need to overly worry about the transferability.

\subsection{Regarding the feasibility of describing nonlinear traffic using linear methods}
Strictly speaking, traffic flow shows nonlinear characteristics.
However,  a number of empirical studies recently showed that the propagation of traffic
along straight lines \cite{Zielke2008,Chen2012i,Laval2014,Jiang2014,He2015a,Li2020a,Dahiyal2020}, and the characteristics could well described using the triangular fundamental diagrams \cite{Newell1982,Cassidy1995,Chiabaut2009,He2017c}.
The TSE methods based on the piecewise linear fundamental diagrams also achieved satisfactory estimation accuracy \cite{Laval2012,Deng2013}. 
Therefore, it is feasible for a (piecewise) linear method to estimate such traffic states. 
Moreover, considering all traffic states in the surrounding cells further improves the capability of reproducing complex relationship, although the relationship between every two cells is assumed to be linear.

\subsection{Regarding multicollinearity of regression}
To avoid multicollinearity, it is common to retain only one of the highly correlated variables in a regression analysis.
However, in the refinement problem of this paper, all variables, i.e., the speeds of the 9 cells in Figure~\ref{fig:Cell}, might be highly correlated due to the spatiotemporal closeness of these cells, and any of them is highly likely to influence details of the refinement.
Therefore, we could not discard any of them to achieve a high prediction accuracy, in particular to capture important traffic patterns, such as the directions of traffic waves.

\subsection{Regarding computational cost}
The proposed model (i.e., Equation~\ref{equ:y} and Table~\ref{tab:Result}) is a multiple linear model and thus high computational efficiency can be guaranteed by the simple and basic formulation of the model.

\section{Conclusion}\label{sec:Conclusion}

To make widely-existed but coarse TS diagrams exhibit more detailed traffic dynamics, this paper introduces a TS diagram refinement problem for the first time and proposes a simple multiple linear regression-based refinement model (Equation~\ref{equ:y} and Table~\ref{tab:Result}) to solve the problem.
The TS diagrams with 6 different cell sizes, i.e., 30s$\times$50m, 60s$\times$100m, 120s$\times$200m, 240s$\times$400m, 30s$\times$200m, and 60s$\times$400m, are selected as examples and considered in modeling.
Three datasets are employed to propose and evaluate the model, i.e., the Ikeda-Route-11 dataset in Japan, the Wangan-Route-4 dataset in Japan, and the US-101 dataset in the United States (Table~\ref{tab:Data}).
In parameter estimation, the proposed model achieves a high goodness of fit with $R^2>0.7$ in the free-flow traffic condition and $R^2>0.9$ in the congested traffic condition (Table~\ref{tab:Result}).

The 1-4 and 1-4-16 tests are then conducted to evaluate the accuracy and transferability of the proposed refinement model.
The 1-4 test refines a TS diagram by increasing the resolution 4 times, and the 1-4-16 test is a stricter test for increasing the resolution 16 times.
Remarkably, the model shows satisfactory accuracy of estimation in the tests:
The ranges of the MAE and MAPE in the 1-4 test are (1.6, 5) km/h and (0.03, 0.1), respectively (Table~\ref{tab:Result14});
The ranges of the MAE and MAPE in the 1-4-16 test are (2.5, 8) km/h and (0.05, 0.17), respectively (Table~\ref{tab:Result146}).
Visual presentations of the refinement results of the two tests (Figures~\ref{fig:STPlot146} and \ref{fig:US101}) are further made to better illustrate the effectiveness of the proposed model.
The source of errors is carefully analyzed, and it is found that some very detailed traffic dynamics that have been omitted from an individual cell in the initial low-resolution TS diagram are difficult to reproduce in a refinement.

Moreover, model comparison demonstrates that the proposed model outperforms the classic ASM in refining TS diagrams and the resulting MAPEs are even less than half of those from the ASM. 
In particular, the good performance of the proposed method in refining the TS diagram whose resolution (i.e., 60s$\times$400m) is specifically selected to align with the typical resolution of stationary detector data demonstrates its capability in handling commonly-existed stationary detector data.


It is believed that the proposed refinement model is rather significant after considering the importance and wide-existence of TS diagrams in transportation research and engineering.
It would be a powerful tool that could quickly and effectively ``clean" coarse TS diagrams and enable them to exhibit ample traffic details.
If needed, one may attempt to refine TS diagrams of traffic flow and density using similar multiple linear regression-based methods; it is highly feasible due to the existence of the fundamental relationship among traffic speed, flow, and density.
Further improving the refinement accuracy, reproducing more omitted details, and making the missed edges complimented will be possible future research directions.
Incorporating more knowledge of traffic flow theory and characteristics may be essential and helpful.
Moreover, deep learning provides us with fruitful and capable tools that might have the potential to make breakthroughs feasible.
However, the strategy for introducing a deep-learning method should be tactical since high-fidelity traffic data that could be used for model training is usually very limited.
Traffic simulation might be an option for generating samples for training and it is worth a try in the future study.


\ifCLASSOPTIONcaptionsoff
  \newpage
\fi

\bibliographystyle{IEEEtran}
\bibliography{library}

\vspace{-5mm}

\begin{IEEEbiography}[{\includegraphics[width=1in,height=1.25in,clip,keepaspectratio]{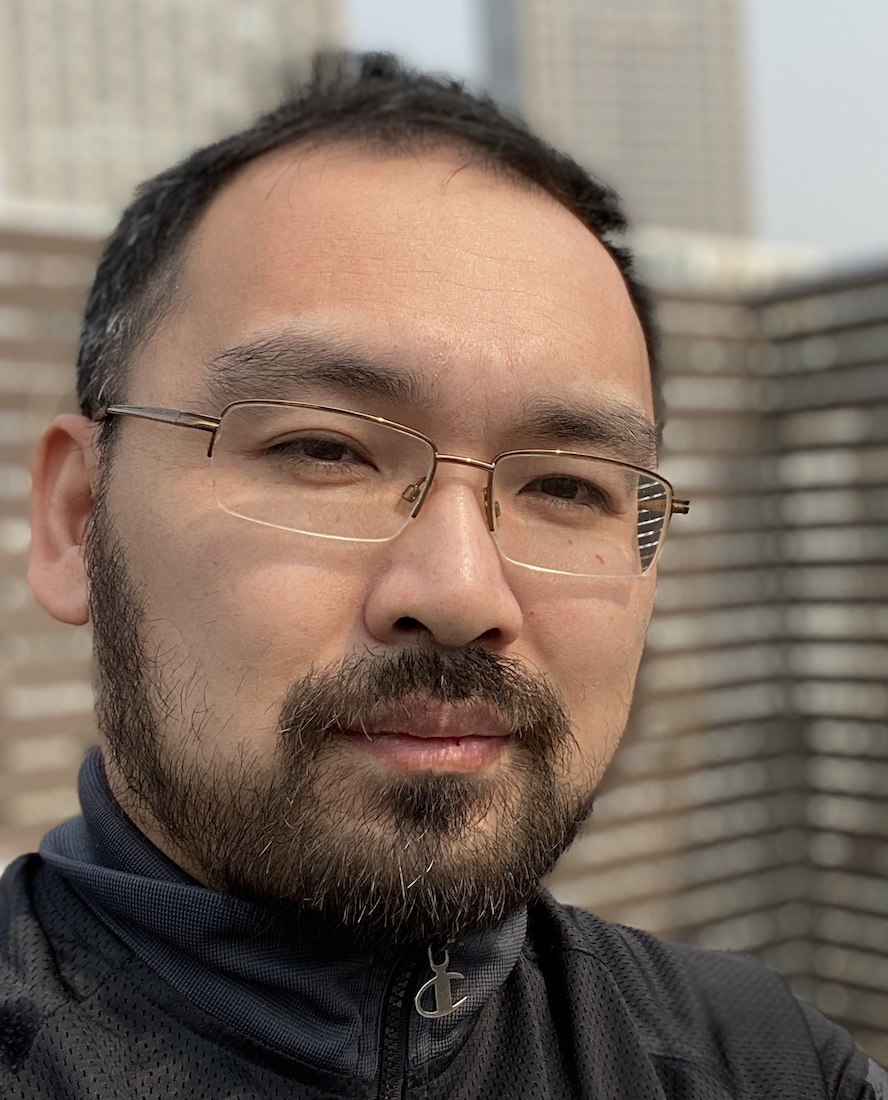}}] 
{Zhengbing He} (M'17-SM'20) received the Bachelor of Arts degree in English language \& literature from Dalian University of Foreign Languages, China, in 2006, and the Ph.D. degree in Systems Engineering from Tianjin University, China, in 2011. 
He was a Postdoctoral Researcher and an Assistant Professor at Beijing Jiaotong University, China. 
From 2018 to 2022, he was a Full Professor at Beijing University of Technology, China. 
Presently, he is a Senior Research Fellow at {\it Senseable City Lab} of Massachusetts Institute of Technology, United States.

His research interests include traffic flow theory, urban mobility, sustainable transportation, etc. 
He has published more than 130 academic papers and 1 book. 
He was listed as {\it World’s Top 2\% Scientists}.
He is the Editor-in-Chief of {\it Journal of Transportation Engineering and Information (Chinese)}. 
Meanwhile, he serves as an Associate Editor for 
{\it IEEE Transactions on Intelligent Transportation Systems}, etc., as a Handling Editor for {\it Transportation Research Record},
and as an Editorial Advisory Board member for {\it Transportation Research Part C}.
He was a Guest Editor for {\it Transportation Research Part C/D}, {\it Journal of Intelligent Transportation Systems}, and {\it Transportmetrica A}, etc. 
He is a member of {\it CAST United Nations Consultative Committee on Transport \& Sustainable Infrastructure}.
His webpage is https://www.GoTrafficGo.com and his email is he.zb@hotmail.com.
\end{IEEEbiography}

\end{document}